\def\year{2021}\relax
\documentclass[letterpaper]{article} % DO NOT CHANGE THIS
\usepackage{aaai21}  % DO NOT CHANGE THIS
\usepackage{times}  % DO NOT CHANGE THIS
\usepackage{helvet} % DO NOT CHANGE THIS
\usepackage{courier}  % DO NOT CHANGE THIS
\usepackage[hyphens]{url}  % DO NOT CHANGE THIS
\usepackage{graphicx} % DO NOT CHANGE THIS
\usepackage{natbib}  % DO NOT CHANGE THIS AND DO NOT ADD ANY OPTIONS TO IT
\usepackage{caption} % DO NOT CHANGE THIS AND DO NOT ADD ANY OPTIONS TO IT
\usepackage{booktabs}
\usepackage{amsmath}
\usepackage{amsfonts}
\usepackage{amsthm}
\usepackage{color}
\newtheorem{theorem}{Theorem}
\usepackage{multirow}
\usepackage[switch]{lineno}  %
\usepackage{enumitem} 
\newcommand{\fixme}[1]{\textcolor{black}{#1}}

\title{
	Topology-Aware Correlations Between Relations for Inductive Link Prediction \\ in Knowledge Graphs
}
\author{Jiajun Chen, Huarui He, Feng Wu, Jie Wang\thanks{Corresponding author.}
	\\
}
\affiliations{
	University of Science and Technology of China\\ 
	\{jiajun98, huaruihe\}@mail.ustc.edu.cn\\
	\{fengwu, jiewangx\}@ustc.edu.cn
}
\begin{document}
	\setcounter{secnumdepth}{1}
	% \linenumbers  %
	\maketitle
	
	\begin{abstract}
		Inductive link prediction---where entities during training and inference stages can be different---has been shown to be promising for completing continuously evolving knowledge graphs. Existing models of inductive reasoning mainly focus on predicting missing links by learning logical rules. However, many existing approaches do not take into account \mbox{\textit{semantic}} \textit{correlations} between relations, which are commonly seen in real-world knowledge graphs. To \mbox{address} this challenge, we propose a novel inductive reasoning approach, namely TACT, which can effectively exploit \textbf{T}opology-\textbf{A}ware \textbf{C}orrela\textbf{T}ions between relations in an entity-independent manner. TACT is inspired by the observation that the semantic correlation between two relations is highly correlated to their topological structure in knowledge graphs. Specifically, we categorize all relation pairs into several \textit{topological patterns}, and then propose a \mbox{Relational} Correlation Network (RCN) to learn the importance of the different patterns for inductive link prediction. Experiments demonstrate that TACT can effectively model semantic \mbox{correlations} between relations, and significantly outperforms existing state-of-the-art methods on benchmark datasets for the inductive link prediction task.
	\end{abstract}
	
	\section{Introduction}
	Knowledge graphs store quantities of structured human knowledge in the form of factual triples, which have been widely used in many fields, such as natural language processing \cite{NLP-field}, recommendation systems \cite{RS-field}, and question answering \mbox{\cite{QA-field}}.
	
	For real-world {knowledge graphs}, new entities keep emerging continuously, such as new users and products in e-commerce {knowledge graphs} and new molecules in biomedical {knowledge graphs} \cite{grail}. Moreover, they usually face the incompleteness problem, i.e., some links are missing. To address this challenge, researchers have paid increasing attention to the {inductive link prediction} task \cite{ ILP-Hai, OOKB, LAN, grail}. Inductive link prediction aims at predicting missing links between entities in knowledge graphs, where entities during training and inference stages can be different. Despite the importance of inductive link prediction in real-world applications, many existing works focus on transductive link prediction and cannot manage previously unseen entities \cite{drum}. Inductive link prediction is challenging as we need to determine which relation it is between two unseen entities during training, that is, we need to generalize what is learned from training entities to unseen entities. 
	
	Existing models of inductive reasoning mainly focus on predicting missing links by learning logical rules in knowledge graphs. Rule learning based methods \cite{neural-lp, drum} explicitly mine logical rules based on observed co-occurrence patterns of relations, which are inherently inductive as the learned rules are entity-independent and can naturally generalize to new entities. More recently, GraIL \cite{grail} implicitly learns logical rules with reasoning over subgraph structures in an entity-independent manner. However, many existing inductive reasoning approaches do not take into account neighboring relational triples when predicting the missing links.
	
\begin{figure}[t]
		\centering 
		\includegraphics[width=0.9\columnwidth]{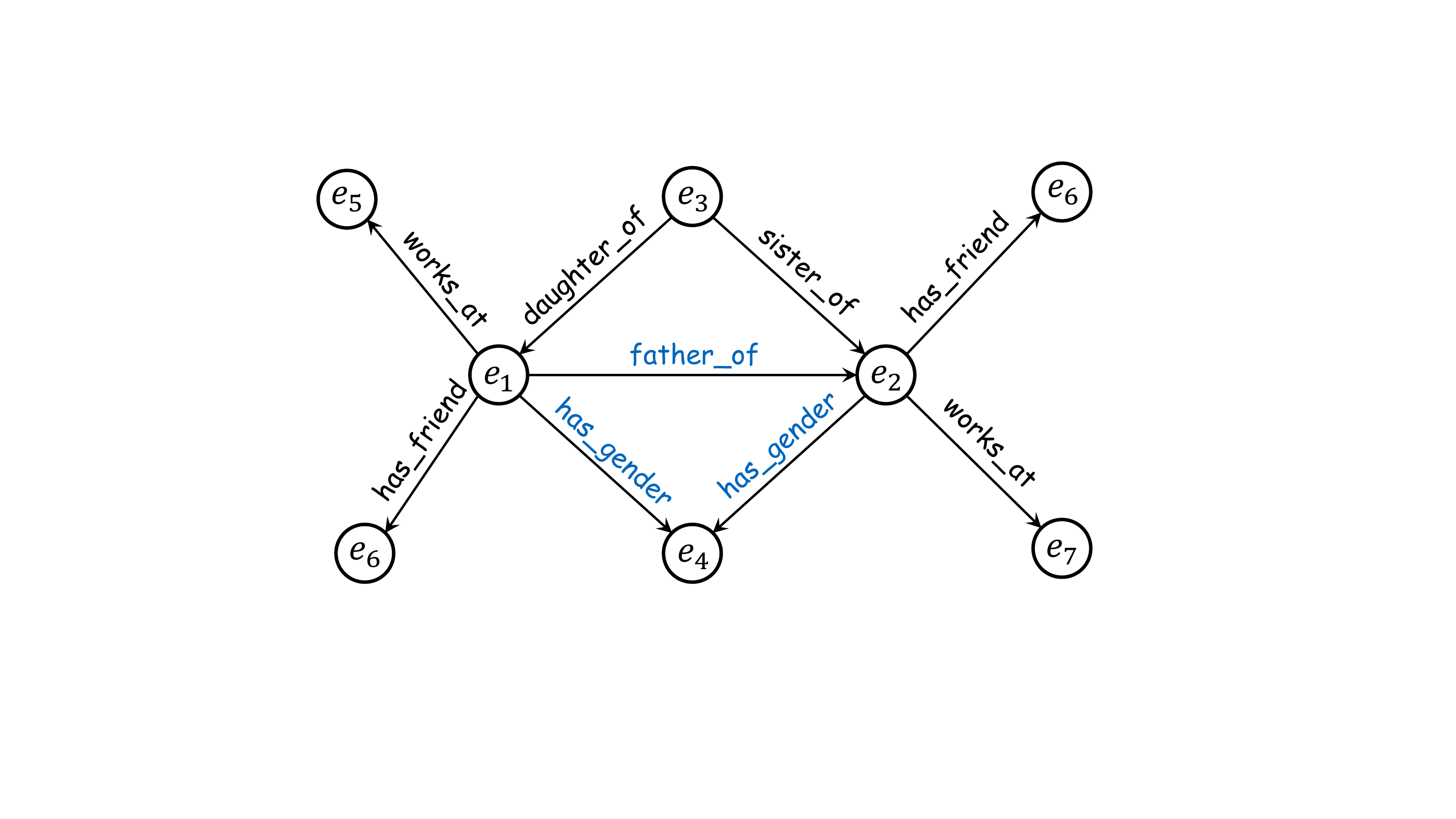}
		\caption{An example in knowledge graphs.}
		\label{fig:1}
	\end{figure}

	To take advantage of the neighboring relational triples, we exploit \textit{semantic correlations} between relations, which are commonly seen in knowledge graphs. For example, for the relations in Freebase \cite{freebase}, ``/people/person/nationality" and ``/people/ethnicity/languages{\_}spoken" have a strong correlation as a person's spoken languages are correlated to the person's nationality, while ``/people/person/nationality" and ``/film/film/country" have a weak correlation. Moreover, the topological patterns between any two relations---the way how they are connected in knowledge graphs---can be different, which influence the correlation patterns. For example, for the relation pair ``father\_of" and ``has\_gender" in Figure \ref{fig:1}, they are connected by $e_1$ in a tail-to-tail manner and connected by $e_2$ in a head-to-tail manner, which are different topological patterns. 
	
	In this paper, we propose a novel inductive reasoning approach, namely TACT, which can effectively exploit \textbf{T}opology-\textbf{A}ware \textbf{C}orrela\textbf{T}ions between relations in knowledge graphs. Specifically, TACT models the semantic correlations in two aspects: correlation patterns and correlation coefficients. We categorize all relation pairs into several different correlation patterns according to their topological structures. We then convert the original knowledge graph to a Relational Correlation Graph (RCG), where the nodes represent the relations and the edges indicate the correlation patterns between any two relations in the original knowledge graph. Based on the RCG, we propose a Relational Correlation Network (RCN) to learn the correlation coefficients of the different patterns for inductive link prediction. TACT can effectively incorporate the information of neighboring relations, thus to promote the performance of link prediction in the inductive setting. Experiments demonstrate that TACT can effectively model semantic correlations between relations, as well as significantly outperforms existing state-of-the-art methods on benchmark datasets for the inductive link prediction task.

	\noindent\textbf{Notations}\hspace{1.5mm}
	Given a set $\mathcal{E}$ of entities and a set $\mathcal{R}$ of relations, a knowledge graph $\mathcal{G}=\{(u,r, v)|u,v\in\mathcal{E},\,r\in\mathcal{R}\}$ is a collection of facts, where $u$, $v$, and $r$ represent head entities, tail entities, and relations between head and tail entities, respectively. We use $\textbf{e}_u, \textbf{r}, \textbf{e}_v$ to denote the embedding of the head entity, the relation and the tail entity. Let $d$ denote the embedding dimension. We denote $i$-th entry of a vector $\textbf{e}$ as $[\textbf{e}]_i$. Let $\circ$ be the Hadamard product between two vectors,
	\begin{align*}
		[\textbf{a} \circ \textbf{b}]_i = [\textbf{a}]_i \cdot [\textbf{b}]_i ,
	\end{align*}
	and we use $\oplus$ to denote the concatenation of vectors.

	\section{Related Work}
	\noindent\textbf{Rule learning based methods}\hspace{1.5mm}
	Rule learning based methods learn logical rules based on observed co-occurrence patterns of relations, which is inherently inductive as the learned rules are independent of entities. Mining rules from data is the central task of inductive logic programming \cite{ILP}. Traditional methods suffer from the problem of scaling to large datasets or being challenging to optimize. More recently, Neural LP \cite{neural-lp} proposes an end-to-end differentiable framework to learn both the structure and parameters of logical rules. DRUM \cite{drum} further improves Neural LP with mining more correct logical rules. However, rule learning based methods mainly focus on mining horn rules, which limits their ability to model more complex semantic correlations between relations in knowledge graphs.
	
	\noindent\textbf{Embedding based methods}\hspace{1.5mm}
	Knowledge graph embedding has been shown to be a promising direction for knowledge graph reasoning \cite{rotate, HAKE, DURA}. Some embedding based methods can generate embeddings for unseen entities. \citet{OOKB} and \citet{LAN} learn to generate entity embeddings for unseen entities by aggregating neighbor entity embeddings with graph neural networks. However, they need new entities to be surrounded by known entities, which cannot handle entirely new graphs. GraIL \cite{grail} develops a graph neural network based link prediction framework that reasons over local subgraph structures, which can perform inductive link prediction in an entity-independent manner. However, GraIL fails to model semantic correlations between relations, which are common in knowledge graphs.
	
	\noindent\textbf{Link prediction using GNNs}\hspace{1.5mm}
	In recent years, graph neural network \cite{gnn-semi, gat} shows great potential in link prediction, as knowledge graphs naturally have graph structures. \citet{RGCN} propose a relational graph neural network to consider the connected relations when applying aggregation on the entities. More recently, \citet{RGHAT} propose a relational graph neural network with hierarchical attention to effectively utilize the neighborhood information of entities in knowledge graphs. However, those methods have difficulty in predicting missing links between unseen entities, as they do link prediction relying on the learned entity embeddings during training.
	
	\noindent\textbf{Modeling correlations between relations}\hspace{1.5mm} Several existing knowledge graph embedding methods consider the problem of modeling correlations between relations. \citet{TransF} decompose the relation–specific projection spaces into a small number of spanning bases, which are shared by all relations. \citet{COR-1} propose to learn the embedded relation matrix by decomposing it as a product of two low-dimensional matrices. Different from the aforementioned work, our work
	\begin{enumerate}[itemindent=0.5em]
		\item[(a)] innovatively categorizes all relation pairs into seven \textit{topological patterns} and propose a novel relational correlation network to model \textit{topology-aware} correlations.
		\item[(b)] considers the inductive link prediction task, while the aforementioned knowledge graph embedding methods have difficulty in dealing with it.
		\item[(c)] outperforms existing state-of-the-art inductive reasoning approaches on benchmark datasets. 
	\end{enumerate}
	
	\begin{figure*}[ht]
		\centering
		\includegraphics[width=1.8\columnwidth]{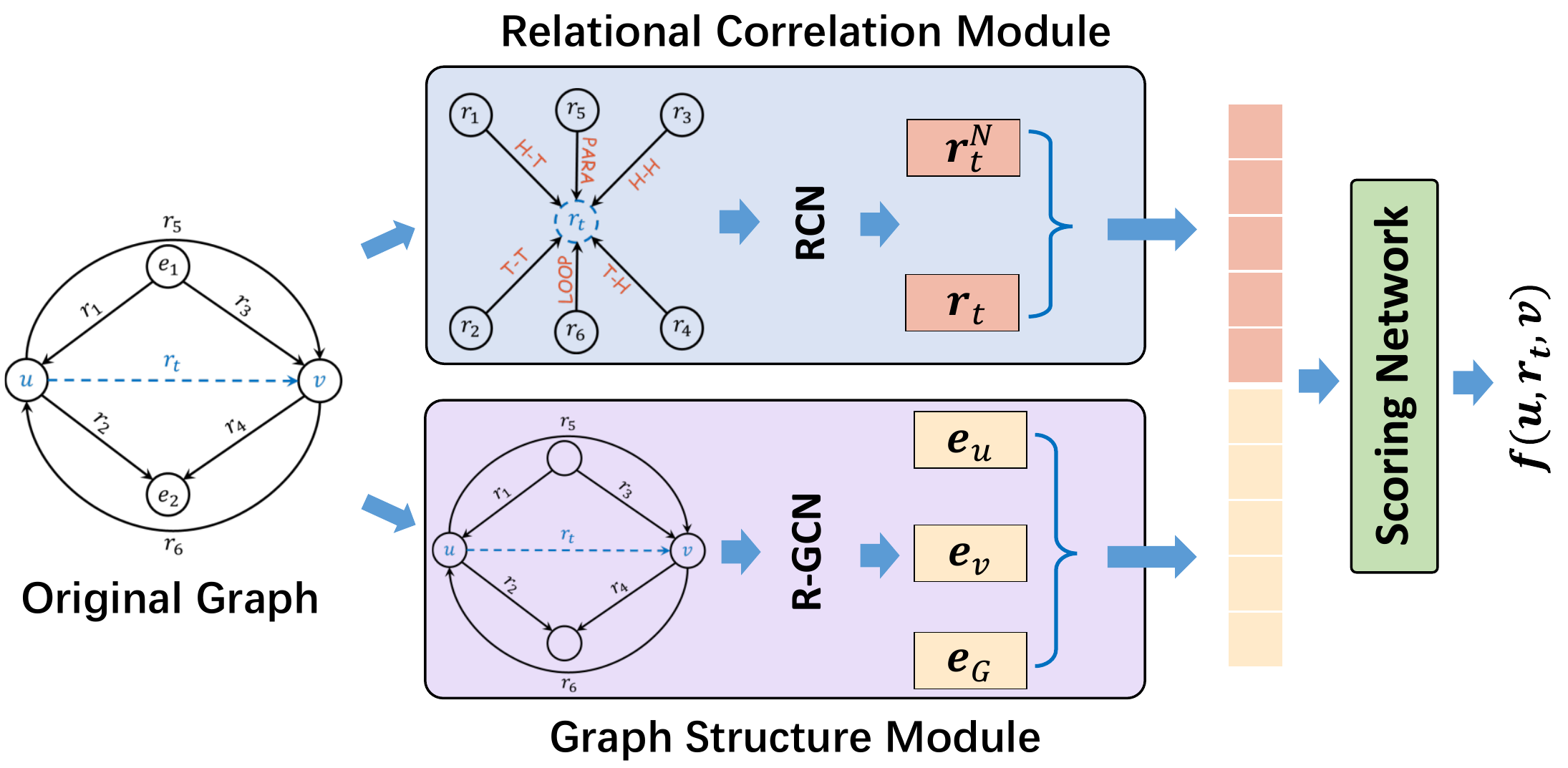}
		\caption{An overview of TACT. TACT consists of two modules: the relational correlation module and the graph structure module. We use a scoring network to score a triple based on the output of the two modules.}
		\label{fig:method-all}
	\end{figure*}

	\section{Methods}
	
	In this section, we introduce our proposed method TACT. To perform inductive link prediction, TACT aims at scoring a given triple $(u, r_t, v)$ in an entity-independent way, where $r_t$ is the target relation between the entity $u$ and $v$. Specifically, TACT consists of two modules: the relational correlation module and the graph structure module. The relational correlation module is proposed based on the observation that semantic correlations between any two relations are highly correlated to their topological structures, which are commonly seen in knowledge graphs. Moreover, we design a graph structure module based on the idea of GraIL \cite{grail} to take advantage of graph structure information. TACT organizes the two modules in a unified framework to perform inductive link prediction. Figure \ref{fig:method-all} gives an overview of TACT.

	\subsection{Modeling correlations between relations}
	To model semantic correlations between relations, we consider the correlations in two aspects: 
	\begin{enumerate}[itemindent=1em]
		\item[(a)] Correlation patterns: The correlations between any two relations are highly correlated to their topological structures in knowledge graphs. 
		\item[(b)] Correlation coefficients: We use correlation coefficients to represent the degree of semantic correlations between any two relations. 
	\end{enumerate}
	
\begin{figure}[ht]
		\centering
		\includegraphics[width=0.75\columnwidth]{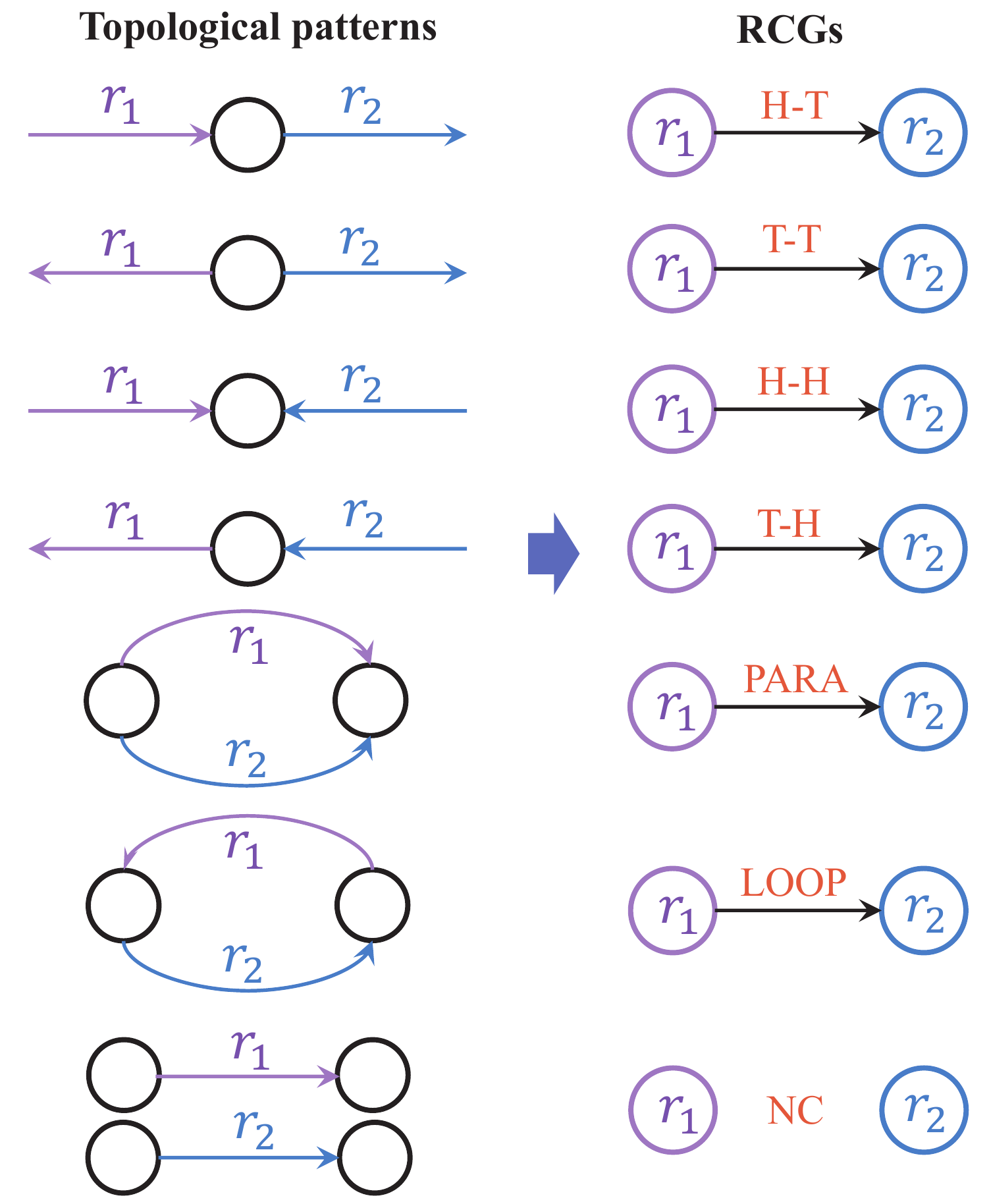}
		\caption{An illustration of the topological patterns between two relations and the corresponding RCGs. For the topological pattern where two relations are not connected, its corresponding RCG consists of two isolated nodes.}
		\label{fig:2}
	\end{figure}

	\noindent\textbf{Relational Correlation Graph}\hspace{1.5mm}
	To model the correlation patterns between any two relations, we categorizes all relation pairs into seven topological patterns. As illustrated in Figure \ref{fig:2}, the topological patterns are ``head-to-tail", ``tail-to-tail", ``head-to-head", ``tail-to-head", ``parallel", ``loop" and ``not connected". We define the corresponding correlation patterns as ``H-T", ``T-T", ``H-H", ``T-H", ``PARA", ``LOOP" and ``NC", respectively. {For example, we denote $(r_1, \text{H-T}, r_2)$ as the correlation between $r_1$ and $r_2$ is the ``H-T" pattern for $r_2$, which indicates that $r_1$ and $r_2$ are connected in a head-to-tail manner.} $(r_1, \text{PARA}, r_2)$ indicates that the two relations {are connected by the same head entity and tail entity}, and $(r_1, \text{LOOP}, r_2)$ indicates that the two relations form a loop in a local graph. We prove that the number of topological patterns between any two relations are at most \textit{seven} in the supplementary.
	
	Based on the definition of different correlation patterns, we can convert the original graph to a Relational Correlation Graph (RCG), where the nodes represent the relations and the edges indicate the correlation patterns between any two relations in the original knowledge graph. Figure \ref{fig:2} shows the topological patterns between any two relations and the corresponding RCGs. Notice that for the topological pattern that two relations are not connected, its corresponding RCG consists of two isolated nodes.
	
	\noindent\textbf{Relational Correlation Network}\hspace{1.5mm}
	Based on the RCG, we propose a Relational Correlation Network (RCN) to model the importance of different correlation patterns for inductive link prediction. RCN consists of two parts: the correlation pattern part and the correlation coefficient part. The correlation pattern part considers the influence of different topological structures between any two relations. Moreover, the correlation coefficient part aims at learning the degree of different correlations.
	
	For an edge with relation $r_t$, we can divide all its adjacent edges in the RCG into six groups by the topological patterns ``H-T", ``T-T", ``H-H", ``T-H", ``PARA" and ``LOOP", respectively. Notice that the topological pattern ``NC" is not considered as it means the edges (relations) are not connected in the knowledge graph. For the six groups, we use six linear transformations to learn the different semantic correlations corresponding to the topological patterns. To differentiate the degree of different correlations for the relation $r_t$, we further use attention networks to learn the correlation coefficients for all the correlations. 
	
	Specifically, we aggregate all the correlations for the relation $r_t$ to get the neighborhood embedding in a local graph, which is denoted by $\textbf{r}_t^{N}$. 
	\begin{align}\label{rel-agg}
		\textbf{r}_t^{N} = \frac{1}{6}\sum_{p=1}^{6} (\textbf{N}^p_t \circ \mathbf{\Lambda}^p_t) \textbf{R} \textbf{W}^p 
	\end{align}
	where $\textbf{W}^p \in \mathbb{R}^{d\times d}$ is the weight parameter, $\textbf{R} \in \mathbb{R}^{|\mathcal{R}|\times d}$ denotes the embedding of all relations. Suppose the embedding of $r_i$ is $\textbf{r}_i \in \mathbb{R}^{1\times d}$, then $\textbf{R}_{[i,:]} = \textbf{r}_i$ where $\textbf{R}_{[i,:]}$ denotes the $i$-th slice along the first dimension. $\textbf{N}^p_t \in \mathbb{R}^{1\times |\mathcal{R}|}$ is the indicator vector where the entry $[\textbf{N}^p_t]_i=1$ if $r_i$ and $r_t$ are connected in the $p$-th topological pattern, otherwise $[\textbf{N}^p_t]_i=0$. $\mathbf{\Lambda}^p_t \in \mathbb{R}^{1\times|\mathcal{R}|}$ is the weight parameter which indicates the degree of different correlations for the relation $r_t$ in the $p$-th correlation pattern. Note that, we restrict $[\mathbf{\Lambda}_t^p]_i \ge 0$ and $\sum_{i=1}^{|\mathcal{R}|} [\mathbf{\Lambda}_t^p]_i = 1$. 
	
	Furthermore, we concatenate $\textbf{r}_t$ and $\textbf{r}_t^{N}$ to get the final embedding $\textbf{r}^F_t$.
	\begin{align}\label{rel-agg-2}
		\textbf{r}^F_t =\sigma ([\textbf{r}_t \oplus \textbf{r}_t^{N}] \textbf{H})
	\end{align}
	where $\textbf{H}\in \mathbb{R}^{2d\times d}$ is the weight parameters, and $\sigma$ is a activation function, such as $\text{ReLU}(\cdot)=\max(0, \cdot)$. We call the module that models semantic correlations between relations as the \textit{relational correlation module}, and $\textbf{r}^F_t$ is the final output of the module.
	
	\begin{table*}[ht]
		\caption{AUC-PR results on the inductive benchmark datasets extracted from WN18RR, FB15k-237 and NELL-995. The results of Neural LP, DURM and GraIL are taken from the paper \cite{grail}.}\label{Tab: AUC-PR}
		\centering
		\resizebox{2.0\columnwidth}{!}{
			\begin{tabular}{l c c c c c  c c c c  c c c c } 
				\toprule
				&\multicolumn{4}{c}{\textbf{WN18RR}}&  \multicolumn{4}{c}{\textbf{FB15k-237}} & \multicolumn{4}{c}{\textbf{NELL-995}}\\
				\cmidrule(lr){2-5} \cmidrule(lr){6-9} \cmidrule(lr){10-13}
				& v1    & v2    & v3    &  v4   & v1    & v2    & v3    &  v4   & v1    & v2    & v3    &  v4 \\
				\midrule
				Neural LP       & 86.02 & 83.78 & 62.90 & 82.06 & 69.64 & 76.55 & 73.95 & 75.74 & 64.66 & 83.61 & 87.58 & 85.69 \\
				DRUM            & 86.02 & 84.05 & 63.20 & 82.06 & 69.71 & 76.44 & 74.03 & 76.20 & 59.86 & 83.99 & 87.71 & 85.94 \\
				% RuleN           & 90.26 & 89.01 & 76.46 & 85.75 & 75.24 & 88.70 & 91.24 & 91.79 & 84.99 & 88.40 & 87.20 & 80.52 \\
				GraIL           & 94.32 & 94.18 & 85.80 & 92.72 & 84.69 & 90.57 & 91.68 & 94.46 & 86.05 & 92.62 & 93.34 & 87.50 \\ \midrule
				TACT-base    & \textbf{98.11} & 97.11 & 88.34 & \textbf{97.25} & 87.36 & \textbf{94.31} & \textbf{97.42} & 98.09 & 94.00 & 94.44 & 93.98 & 94.93 \\
				TACT  & 96.15 & \textbf{97.95} & \textbf{90.58} & 96.15 & \textbf{88.73} &  94.20 & 97.10 & \textbf{98.30} & \textbf{94.87}& \textbf{96.58} & \textbf{95.70} & \textbf{96.12} \\
				\bottomrule
			\end{tabular}
		}
	\end{table*}
	
	\subsection{Modeling graph structures}
	For a triple $(u, r_t, v)$, {the local graph around it} contains the information about how the triple connected with its neighborhoods. To take advantage of the local graph structural information, we use a graph structural network to embed local graphs into vectors based on GraIL \cite{grail}. To model {the graph structure around the triple $(u, r_t, v)$,} we perform the following steps: (1) subgraph extraction; (2) node labeling; (3) graph embedding. Notice that nodes represent entities in knowledge graphs.
	
	\noindent\textbf{Subgraph extraction}\hspace{1.5mm}
	For a triple $(u, r_t, v)$, we first extract the enclosing subgraph around the target nodes $u$ and $v$ \cite{grail}. The enclosing subgraph between nodes $u$ and $v$ is given by the following steps. First, we compute the neighbors $\mathcal{N}_k(u)$ and $\mathcal{N}_k(v)$ of the two nodes $u$ and $v$, respectively, where $k$ denotes the max distance of neighbors. Second, we take an intersection of $\mathcal{N}_k(u)$ and $\mathcal{N}_k(v)$ to get $\mathcal{N}_k(u) \cap \mathcal{N}_k(v)$. Third, we compute the enclosing subgraph $\mathcal{G}(u, r_t, v)$ by pruning nodes of $\mathcal{N}_k(u) \cap \mathcal{N}_k(v)$ that are isolated or at a distance larger than $k$ from either node $u$ or $v$.

	\noindent\textbf{Node labeling}\hspace{1.5mm}
	Afterwards, we label the nodes in the extracted enclosing subgraph following \cite{grail}. We label each node $i$ in the subgraph around nodes $u$ and $v$ with the tuple $(d(i,u), d(i,v))$, where $d(i,u)$ denotes the shortest distance between nodes $i$ and $u$ without counting any path through $v$ (likewise for $d(i,v)$). This captures the topological position of each node with respect to the target nodes. The two target nodes $u$ and $v$ are uniquely labeled $(0,1)$ and $(1,0)$. The node features are defined as $[\text{one-hot}(d(i,u)) \oplus \text{one-hot}(d(i,v))]$, where $\text{one-hot}(p) \in \mathbb{R}^{1\times d}$ represent the one-hot vector that only the $p$-th entry is $1$, where $d$ represent the dimension of embeddings.

	\noindent\textbf{Graph embedding}\hspace{1.5mm}
	After node labeling of the enclosing subgraph, the nodes in the subgraph have initial embeddings. We use R-GCN \cite{RGCN} to learn the embeddings on the extracted enclosing subgraph $\mathcal{G}(u, r_t, v)$.
	\begin{align*}
		\textbf{e}_i^{(k+1)} = \sigma\left( \sum_{r\in \mathcal{R}} \sum_{j\in \mathcal{N}_i^r} \frac{1}{c_{i,r}} \textbf{e}_j^{(k)}  \textbf{W}_r^{(k)} + \textbf{e}_i^{(k)}\textbf{W}_0^{(k)} \right)
	\end{align*}
	where $\textbf{e}_i^{(k)}$ denotes the embedding of entity $e_i$ of the $k$-th layer in the R-GCN. $\mathcal{N}_i^r$ denotes the set of neighbor indices of node $i$ under relation $r\in \mathcal{R}$. $c_{i,r} = |\mathcal{N}_i^r|$ is a normalization constant. $\textbf{W}_r^{(k)}\in\mathbb{R}^{d\times d} (r\in\mathcal{R}), \textbf{W}_0^{(k)} \in \mathbb{R}^{d\times d}$ are the weight parameters. $\sigma(\cdot)$ is a activation function, such as the $\text{ReLU}(\cdot) = \max(0, \cdot)$.
	
	Suppose that the number of layers in R-GCN is $L$, we calculate the embedding of the subgraph $\mathcal{G}(u,r_t,v)$ as
	\begin{align*}
		\textbf{e}_{\mathcal{G}(u,r_t,v)}^{(L)} = \frac{1}{|\mathcal{V}_{\mathcal{G}(u,r_t,v)}|} \sum_{i\in \mathcal{V}_{\mathcal{G}(u,r_t,v)}} \textbf{e}_i^{(L)},
	\end{align*}
	where $\mathcal{V}_{\mathcal{G}(u,r_t,v)}$ denotes the set of nodes in graph $\mathcal{G}(u,r_t,v)$. Combining the target nodes and subgraph embedding, the structural information is represented by the vector $\textbf{e}_S \in \mathbb{R}^{1\times 3d}$,
	\begin{align*}
		\textbf{e}_S = \textbf{e}_{\mathcal{G}(u,r_t,v)}^{(L)} \oplus \textbf{e}_u^{(L)} \oplus \textbf{e}_v^{(L)}
	\end{align*}
	We call the module that models graph structures as the \textit{graph structure module}, and $\textbf{e}_S$ is the final output of the module.

	\subsection{The framework of TACT}
	\noindent\textbf{Scoring network}\hspace{1.5mm}
	The relational correlation module and graph structure module output the embedding vectors $\textbf{r}_t^F$ and $\textbf{e}_S$, respectively. To organize the two modules in a unified framework, we design a scoring network to combine the outputs of the two modules and get the score for a given triple $(u, r_t, v)$. The score function $f(u, r_t, v)$ is defined as
	\begin{align*}
		f(u,r_t,v) = [\textbf{r}_t^{F} \oplus \textbf{e}_S] \textbf{W}_S
	\end{align*}
	where $\textbf{W}_S \in \mathbb{R}^{4d\times 1}$ is a weight parameter.
	
	\noindent\textbf{Loss function}\hspace{1.5mm}
	We perform negative sampling and train the model to score positive triples higher than the negative triples using a {noise-contrastive} hinge loss following \cite{transe}. The loss function $\mathcal{L}$ is 
	\begin{align*}
		\mathcal{L} = \sum_{i\in [n], (u,r_t,v)\in \mathcal{G}} \max(0, f(u'_i,r'_{t,i},v'_{i}) - f(u,r_t,v) + \gamma)
	\end{align*}
	\noindent where $\gamma$ is the margin hyperparameter and $\mathcal{G}$ denotes the set of all triples in the knowledge graph. $(u'_i,r'_{t,i},v'_{i})$ denotes the $i$-th negative triple of the ground-truth triple $(u, r_t, v)$ and $[n]$ represent the set $\{1,2,\cdots, n\}$, where $n$ is the number of negative samples for each triple.

	\section{Experiments and Analysis}
	This section is organized as follows. First, we introduce the experimental configurations, including datasets, implementation details, and the baseline model. Second, we show the effectiveness of our proposed approach TACT on several benchmark datasets. Finally, we show the results of ablation studies, case studies, and further experiments.  The code of TACT is available on GitHub at \url{https://github.com/MIRALab-USTC/KG-TACT}.
	
	\subsection{Experimental Configurations}
	\noindent \textbf{Datasets}\hspace{1.5mm} We use the benchmark datasets for inductive link prediction proposed in GraIL \cite{grail}, which are derived from WN18RR \cite{wn18rr}, FB15k-237 \cite{conve}, and NELL-995 \cite{xiong2017deeppath}. For inductive link prediction, the train set and the test set should have no overlapping entities. Each knowledge graph of WN18RR, FB15k-237, and NELL-995 induces four versions of inductive datasets with increasing sizes. Details of the datasets are summarized in Table \ref{Tab:ind-data}.

	\begin{table}[ht]
		\caption{Statistics of inductive benchmarks. We use \#E and \#R and \#TR to denote the number of entities, relations, and triples, respectively.}
		\centering
		\resizebox{1.0\columnwidth}{!}{
			\begin{tabular}{l l *{9}{c}}
				\toprule
				& &\multicolumn{3}{c}{\textbf{WN18RR}}&  \multicolumn{3}{c}{\textbf{FB15k-237}} & \multicolumn{3}{c}{\textbf{NELL-995}}\\
				\cmidrule(lr){3-5}\cmidrule(lr){6-8}\cmidrule(lr){9-11}
				& &\#R &\#E &\#TR     &\#R &\#E &\#TR    &\#R &\#E &\#TR   \\
				\midrule
				\multirow{2}{*}{v1} & train & 9 & 2746 & 6678 &   183 & 2000 & 5226 &  14 & 10915 & 5540\\
				& test  & 9 & 922 & 1991 &   146 & 1500 & 2404 &  14 & 225 & 1034\\
				\midrule
				\multirow{2}{*}{v2} & train & 10 & 6954 & 18968 &  203 & 3000 & 12085 &  88 & 2564 & 10109\\
				& test  & 10 & 2923 & 4863 &   176 & 2000 & 5092 &  79 & 4937 & 5521\\
				\midrule
				\multirow{2}{*}{v3} & train & 11 & 12078 & 32150 & 218 & 4000 & 22394 &  142 & 4647 & 20117\\
				& test  & 11 & 5084 & 7470 &   187 & 3000 & 9137 &  122 & 4921 & 9668\\
				\midrule
				\multirow{2}{*}{v4} & train & 9 & 3861 & 9842 &   222 & 5000 & 33916 &  77 & 2092 & 9289\\
				& test  & 9 & 7208 & 15157 &   204 & 3500 & 14554 &  61 & 3294 & 8520\\
				\bottomrule
			\end{tabular}
		}
		\label{Tab:ind-data}
	\end{table}
	
	\noindent\textbf{Training protocol} \hspace{1.5 mm} We use Adam optimizer \cite{adam} with initial learning rate of 0.01 and batch size of 16. We randomly sample two-hop enclosing subgraphs for each triple when training and testing, and use a two-layer GCN to calculate the embeddings of subgraphs. The margins in the loss functions are set to 8, 16, 10 for WN18RR, FB15k-237, NELL-995, respectively. The maximum number of training epochs is set to 10.
	
	\noindent\textbf{The baseline model}\hspace{1.5mm} To evaluate the effectiveness of our proposed relational correlation module, we propose a baseline called TACT-base, which scores a triple $(u, r_t, v)$ only relying on the output of the relational correlation module. That is, the score function of TACT-base is 
	\begin{align*}
		f_{\text{base}}(u, r_t, v) = \textbf{r}_t^{F} \textbf{W}_{base}
	\end{align*}
	where $\textbf{W}_{\text{base}}\in \mathbb{R}^{d\times 1}$ is a weight parameter.
	
	\begin{table*}[ht]
		\caption{MRR and H@1 results on the inductive benchmark datasets extracted from WN18RR, FB15k-237, and NELL-995. We reimplement the three baselines Neural LP, DRUM, and GraIL under the inductive relation prediction protocol, with all the hyperparameters keeping the same with their original papers for a fair comparison. 
			% Notice that, the inductive relation prediction in here is specifically refer to given the head and tail entities to predict the target relation between them in the inductive setting.
		}\label{Tab: MRR}
		\centering
		\resizebox{2.0\columnwidth}{!}{
			\begin{tabular}{l c c c c c  c c c c  c c c c } 
				\toprule
				&\multicolumn{4}{c}{\textbf{WN18RR}}&  \multicolumn{4}{c}{\textbf{FB15k-237}} & \multicolumn{4}{c}{\textbf{NELL-995}}\\
				\cmidrule(lr){2-5} \cmidrule(lr){6-9} \cmidrule(lr){10-13}
				&\multicolumn{2}{c}{v1}& \multicolumn{2}{c}{v4}& \multicolumn{2}{c}{v1}& \multicolumn{2}{c}{v4}& \multicolumn{2}{c}{v1}& \multicolumn{2}{c}{v4}\\
				\cmidrule(lr){2-3}\cmidrule(lr){4-5}
				\cmidrule(lr){6-7}\cmidrule(lr){8-9}
				\cmidrule(lr){10-11}\cmidrule(lr){12-13}
				& MRR   & H@1 & MRR   & H@1 & MRR   & H@1 & MRR   & H@1  & MRR   & H@1 & MRR   & H@1 \\
				\midrule
				Neural LP & .615 & .548 & .351 & .195 & .086 & .073 & .057 & .041 & .175 & .050 & .087 & .032 \\
				DRUM      & .450 & .277 & .429 & .260 & .080 & .054 & .049 & .026 & .238 & .170 & .092 & .018 \\
				GraIL     & .851 & .749 & .746 & .610 & .054 & .016 & .056 & .017 & .337 &	.146 & .072 &	.017 \\
				\midrule
				TACT-base & .990 & .983 & .981 & .966 & .804 & .700 & \textbf{.593} & \textbf{.409} & .877 & .756 & .304 & .171 \\
				TACT    & \textbf{.995} & \textbf{.995} & \textbf{.988} & \textbf{.982} & \textbf{.830} & \textbf{.741} & .575 & .378 & \textbf{.880} & \textbf{.776} & \textbf{.571} & \textbf{.444}  \\
				\bottomrule
			\end{tabular}
		}
		
	\end{table*}
	
	\begin{table*}[ht]
		\caption{Ablation results  the inductive benchmark datasets extracted from WN18RR, FB15k-237, and NELL-995. ``TACT w/o RA" represent the baseline that omits the relation aggregation in TACT. ``TACT w/o RC" represent the baseline that performs relation aggregation without modeling correlations between relations in TACT.}\label{Tab: ablation-1}
		\centering
		\resizebox{2.0\columnwidth}{!}{
			\begin{tabular}{l c c c c c  c c c c  c c c c } 
				\toprule
				&\multicolumn{4}{c}{\textbf{WN18RR}}&  \multicolumn{4}{c}{\textbf{FB15k-237}} & \multicolumn{4}{c}{\textbf{NELL-995}}\\
				\cmidrule(lr){2-5} \cmidrule(lr){6-9} \cmidrule(lr){10-13}
				&\multicolumn{2}{c}{v1}& \multicolumn{2}{c}{v4}& \multicolumn{2}{c}{v1}& \multicolumn{2}{c}{v4}& \multicolumn{2}{c}{v1}& \multicolumn{2}{c}{v4}\\
				\cmidrule(lr){2-3}\cmidrule(lr){4-5}
				\cmidrule(lr){6-7}\cmidrule(lr){8-9}
				\cmidrule(lr){10-11}\cmidrule(lr){12-13}
				& MRR   & H@1 & MRR   & H@1 & MRR   & H@1 & MRR   & H@1  & MRR   & H@1 & MRR   & H@1 \\
				\midrule
				
				TACT w/o RA & .875 & .793 & .806  & .670 & .118 & .041  & .112 & .023 & .473 & .322 & .071 & .023 \\
				TACT w/o RC & .953 & .913 & .965 & .935 & .666 & .539 & .525 & .335 & .669 & .390 & .508 & .317 \\ 
				\midrule
				TACT & \textbf{.995} & \textbf{.995} & \textbf{.988} &	\textbf{.982} & \textbf{.830} & \textbf{.741} & \textbf{.575} & \textbf{.378} & \textbf{.880} & \textbf{.776} & \textbf{.571} & \textbf{.444} \\
				\bottomrule
			\end{tabular}
		}
	\end{table*}
	
	\subsection{Inductive Link Prediction}
	We evaluate the models on both classification and ranking metrics. For both metrics, we compare our method to several state-of-the-art methods, including Neural LP \cite{neural-lp}, DRUM \cite{drum}, and GraIL \cite{grail}.

	\noindent\textbf{Classification metric}\hspace{1.5mm} We use area under the precision-recall curve (AUC-PR) as the classification metric following GraIL \cite{grail}. We replace the head or tail of every test triple with a random entity to sample the corresponding negative triple. Then we score the positive triples with an equal number of negative triples to calculate AUC-PR following GraIL \cite{grail}. We run each experiment five times with different random seeds and report the mean results.
	
	Table \ref{Tab: AUC-PR} shows the AUC-PR results for inductive link prediction. Our baseline model TACT-base outperforms the inductive baselines on all the datasets. As TACT-base totally relies on the relational correlation module to perform link prediction, the results demonstrate the effectiveness of our proposed model for inductive link prediction. TACT further improves the performance of TACT-base and get around $4\%$ improvement against GraIL on most datasets. The experiments demonstrate the effectiveness of modeling topology-aware correlations between relations in TACT for the inductive link prediction task. 
	
	\noindent\textbf{Ranking metric}\hspace{1.5mm} 
	We further evaluate the model for the inductive \textit{relation} prediction to verify the effectiveness of modeling relational correlations in TACT. Inductive relation prediction aims at predicting the target relation between the given head and tail entities. Specifically, for a given relation prediction $(u, ?, v)$ in the test set, we rank the ground-truth relation $r$ against all other candidate relations. Following the standard procedure in prior work \cite{transe}, we use the filtered setting, which does not take any existing valid triples into account at ranking. We choose Mean Reciprocal Rank (MRR) and Hits at N (H@N) as the evaluation metrics. As the baselines are evaluated in the setting of head/tail prediction, we reimplement Neural LP \cite{neural-lp}, DRUM \cite{drum}, and GraIL \cite{grail} under our ranking setting. For a fair comparison, we keep the hyperparameters the same with their original papers. Following GraIL, we run each experiment five times with different random seeds and report the mean results.
	
	Table \ref{Tab: MRR} shows the results of MRR and H@1 on v1 and v4 of WN18RR, FB15k-237, and NELL-995. More results about v2 and v3 of WN18RR, FB15k-237, and NELL-995 are listed in the supplementary. As we can see, TACT significantly outperforms rule learning based methods \cite{neural-lp, drum} and GraIL \cite{grail} for the inductive relation prediction. The improvements on FB15k-237 and NELL-995 are more significant than WN18RR. FB15k-237 and NELL-995 contain much more relations than WN18RR, thus semantic correlations between relations are more complex in FB15k-237 and NELL-995. The experiments show that GraIL has difficulty in modeling relational semantics in the condition that the number of relations is large. In contrast, TACT can model the complex patterns of relations through exploiting correlations between relations in knowledge graphs. TACT-base also can significantly outperform existing state-of-the-art methods.

	\subsection{Ablation Studies}
	In this part, we conduct ablation studies on the relation aggregation and modeling correlations between relations.
	
	\noindent \textbf{TACT w/o RA}\hspace{1.5mm} 
	In our proposed method, we aggregate the relation embedding $\textbf{r}_t$ and neighborhood embedding $\textbf{r}_t^N$ to get the final relation embedding $\textbf{r}_t^F$. We omit the aggregation of neighborhood embedding, that is, we let the output of the relational correlation module to be $\textbf{r}_t$. We called this method ``TACT w/o RA".
	
	\noindent \textbf{TACT w/o RC}\hspace{1.5mm}
	Modeling topology-aware correlations between relations is one of our main contributions. We design a baseline that performs relation aggregation without modeling correlations between relations. That is, the baseline reformulates the equation \eqref{rel-agg} as 
	\begin{align*}
		\textbf{r}_t^{N} = \frac{1}{|\mathcal{N}(r_t)|}\sum_{i\in \mathcal{N}(r_t)} \textbf{r}_i
	\end{align*}
	where $\mathcal{N}(r_t)$ represents the set of neighborhood relations of $r_t$. We call this baseline ``TACT w/o RC" for short.
	
	Table \ref{Tab: ablation-1} shows the results on three benchmark datasets. The experiments demonstrate the effectiveness of modeling topology-aware correlations between relations in TACT. As correlations between relations are common in knowledge graphs, the relation aggregation in ``TACT w/o RC" can take advantage of the neighborhood relations, which is helpful for inductive link prediction. Our proposed method further distinguishes the correlation patterns and correlation coefficients between relations, which makes the learned embeddings of relations more expressive for inductive link prediction. As we can see, TACT significantly outperforms ``TACT w/o RA" and ``TACT w/o RC" on all the datasets. 
	
	\noindent \textbf{The Frequency-based Method}\hspace{1.5mm}
	We conduct an experiment by ranking relations according to their frequencies and compare TACT with the frequency-based method on the datasets. For the frequency-based method, the returned rank list for every prediction is the same, which is the rank according to the relation frequencies from high to low in the KG. In other words, the frequency-based method represent a type of data bias in the datasets. As illustrated in Table \ref{tab:mrr-frequency}, TACT outperforms the frequency-based method by a large margin on the datasets. The results demonstrate that the effectiveness of TACT is not due to the data bias of relation frequencies in benchmark datasets.

	\begin{table}[h]
		\caption{The MRR results of the frequency-based method and our proposed TACT.}
		\begin{center}
			\resizebox{1.0\columnwidth}{!}{
				\begin{tabular}{c c c c}
					\toprule
					&  WN18RR(v1) & NELL-995(v1) & FB15k-237(v1) \\
					\midrule
					frequency-based & 0.763 & 0.201 & 0.470 \\
					TACT & \textbf{0.995} & \textbf{0.830} & \textbf{0.880} \\
					\bottomrule
			\end{tabular}}
			\label{tab:mrr-frequency}    
		\end{center}
	\end{table}

	\begin{table}[ht]
		\caption{Some relations and their top 3 relevant relations. The relations are taken from WN18RR and NELL-995. We use CP to represent correlation pattern, and use CC to represent correlation coefficient.}
		\resizebox{1.0\columnwidth}{!}{
			\begin{tabular}{l c c c}
				\toprule
				Target relation  &  Most relevant relations & CP & CC \\
				\midrule
				
				&\textit{\_has\_part}     & PARA                  &0.68 \\
				\textit{\_member\_meronym} &\textit{\_similar\_to} & H-H          &0.39 \\
				&\textit{\_synset\_domain\_topic\_of} & T-H     &0.31 \\
				\midrule  
				&\textit{\_similar\_to}   & LOOP                 &0.40 \\
				\textit{\_similar\_to} & \textit{\_member\_meronym} & H-H        &0.39 \\
				&\textit{\_instance\_hypernym} & T-T           &0.35 \\
				\midrule
				& \textit{television\_station\_affiliated\_with} & H-H &0.52 \\
				\textit{head\_quartered\_in} & \textit{head\_quartered\_in} & PARA &0.33 \\
				& \textit{acquired}       & T-H                 &0.30 \\
				\bottomrule
		\end{tabular}}
		\label{tab:ex2}
	\end{table}

	\begin{table}[ht]
		\caption{The results for inductive link prediction on the dataset YAGO3-10.}
		\begin{center}
			\resizebox{0.7\columnwidth}{!}{
				\begin{tabular}{c c c c}
					\toprule
					&  AUC-PR & MRR & Hits@1 \\
					\midrule
					GraIL & 0.634 & 0.158 & 0.048 \\
					TACT & \textbf{0.915} & \textbf{0.406} & \textbf{0.140} \\
					\bottomrule
			\end{tabular}}
			\label{tab:yago3}    
		\end{center}
	\end{table}

	% \noindent\textbf{Further Experiments}\hspace{1.5mm}
	\subsection{Further Experiments}
	
	\subsubsection{Case Studies}
	
	We select some relations and show the top three relevant relations of them in table \ref{tab:ex2}. Recall that the sum of correlation coefficients for each correlation pattern is 1. The results show that TACT can learn some correct correlation patterns. For example, among all the neighboring relations of ``{\_member\_meronym}", ``{\_has\_part}"---which is adjacent to ``{\_member\_meronym}" in the topological pattern of parallel---gets the largest correlation coefficient, as ``{\_has\_part}" and ``{\_member\_meronym}" have similar semantics. We list more examples in the \fixme{supplementary material}. 
	
	\subsubsection{Results on YAGO3-10}
	
	To demonstrate the effectiveness of our proposed method on larger KG with few relations. We conduct experiments on YAGO3-10, which is a subset of YAGO3 \citep{yago3} and contains 37 relaitons and 123,182 entities. Table \ref{tab:yago3} show the results for inductive link prediction of GraIL and TACT on YAGO3-10. As we can see, TACT outperforms GraIL by all the metrics, which demonstrate our proposed TACT can effectively deal with larger KG with few relations.
	
	\subsubsection{Running Time} 
	
	Table \ref{tab:running_time} show the running time of GraIL and TACT. TACT would take more time than GraIL for modeling the correlations between relations, but the total running time is still acceptable.

	\begin{table}[ht]
		\caption{The running time of TACT on the datasets.}
		\begin{center}
			\resizebox{1.0\columnwidth}{!}{
				\begin{tabular}{c c c c}
					\toprule
					&  WN18RR(v1) & NELL-995(v1) & FB15k-237(v1) \\
					\midrule
					GraIL & 0.18 h & 0.17 h & 0.32 h \\
					TACT & 0.22 h & 2.53 h & 8.97 h \\
					\bottomrule
			\end{tabular}}
			\label{tab:running_time}    
		\end{center}
	\end{table}
	
	\section{Conclusion}
	In this paper, we propose a novel inductive reasoning approach called TACT, which can effectively exploit topology-aware correlations between relations for inductive link prediction in knowledge graphs. TACT categorizes all relation pairs into several \textit{topological patterns}, and then use the proposed RCN to learn the importance of the different patterns for inductive link prediction. Experiments demonstrate that our proposed TACT significantly outperforms several existing state-of-the-art methods on benchmark datasets for the inductive link prediction task. 
	
	\section*{Acknowledgments}
	We would like to thank all the anonymous reviewers for their insightful comments. This work was supported in part by NSFC ($61822604$, $61836006$, $62021001$, $\text{U}19\text{B}2026$). 
	
	\bibliographystyle{aaai}
	\bibliography{AAAI2021}
	
	\newpage
	\section*{Appendix}

	\section*{A.\,\,\, The Number of Topological Patterns}
	\begin{theorem}
		In the knowledge graph, the number of topological patterns between any two irreflexive relations is at most seven.
	\end{theorem}
	\begin{proof}
		For any two edges $(h_1, t_1)$ and $(h_2, t_2)$ in the knowledge graph, the number of intersections between them can be $0,1,2$. Suppose the two edges are corresponding to two triples $(h_1, r_1, t_1)$ and $(h_2, r_2, t_2)$ in the knowledge graph. As the relations $r_1$ and $r_2$ both are irreflexive, we have $h_1\ne t_1$ and $h_2 \ne t_2$.
		\begin{enumerate}[itemindent=0.5em]
			\item[(a)] If the number of intersections is $0$, we know the two edges are not connected, which implies that their topological structure only has one pattern. 
			\item[(b)] If the number of intersections is $1$, there are four cases: 
			\begin{enumerate}[itemindent=0.5em]
				\item[1)] $h_1=h_2, t_1\ne t_2$;
				\item[2)] $h_1=t_2, h_2\ne t_1$;
				\item[3)] $t_1=h_2, h_1\ne t_2$;
				\item[4)] $t_1=t_2, h_1\ne h_2$.
			\end{enumerate}
			Each case is corresponding to a specific topological pattern. The topological structures can be head-to-head, head-to-tail, tail-to-head and tail-to-tail, thus the number of topological patterns is four.
			\item[(c)] If the number of intersections is $2$, there are two cases:
			\begin{enumerate}[itemindent=0.5em]
				\item[1)] $h_1=h_2, t_1=t_2$;
				\item[2)] $h_1=t_2, h_2=t_1$.
			\end{enumerate}
			Each case is corresponding to a specific topological pattern. The number of topological patterns is two.
		\end{enumerate}
		Therefore, the number of topological patterns between any two irreflexive relations is at most $1+4+2=7$.
	\end{proof}

	\section*{B.\,\,\, Additional Experiments}
	\textbf{Inductive Link Prediction}\,\, Table \ref{Tab: MRR} shows the results of MRR and H@1 on v2 and v3 of WN18RR, FB15k-237, and NELL-995 under the inductive relation prediction protocol. As we can see, TACT significantly outperforms rule learning based methods and GraIL for the inductive relation prediction. 
	TACT achieves more significant improvements on FB15k-237 and NELL-995 than those on WN18RR.
	% The improvements on FB15k-237 and NELL-995 are more significant than WN18RR. 
	Because FB15k-237 and NELL-995 contain much more relations than WN18RR, semantic correlations between relations are more complex in FB15k-237 and NELL-995. The experiments show that GraIL has difficulty in modeling relational semantics in the condition that the number of relations is large. In contrast, TACT can model the complex patterns of relations by exploiting correlations between relations in knowledge graphs. TACT-base also can significantly outperform existing state-of-the-art methods.

	\textbf{The Input of The Scoring Network}\,\, In our method, the input of the scoring network is 
	\begin{align*}
		\textbf{r}_t^{F} \oplus \textbf{e}_{\mathcal{G}(u,r_t,v)}^{(L)} \oplus \textbf{e}_u^{(L)} \oplus \textbf{e}_v^{(L)}     
	\end{align*}
	That is, the score of a triple is related to the final relation embedding $\textbf{r}_t^{F}$, the graph embedding $\textbf{e}_{\mathcal{G}(u,r_t,v)}^{(L)}$, and the node embedding $\textbf{e}_u^{(L)} \oplus \textbf{e}_v^{(L)}$. We conduct some experiments to get access to the effect of each part of embeddings in inductive link prediction. Table \ref{Tab: ablation-2} shows the results of scoring a triple based on different combinations of embeddings. We can see that any part of the embeddings has its own effect on the final results. When performing inductive relation prediction, using the final relation embedding solely---which is exactly the method of our proposed baseline model TACT-base---can get a fairly good performance. This shows that modeling semantic correlation between relations is benificial to make the right relation prediction in the inductive setting. The node embeddings and graph embedding will further promote the performance on different datasets. For example, adding node embeddings will improve the performance on WN18RR. We can set the used embedding parts as a hyperparameter to further improve the performance of TACT to get the best results on different benchmark datasets for inductive link prediction. 
	
	\textbf{Case Studies} \,\, We present some examples of inductive link prediction on the test set to show the effectiveness of TACT in modeling correlations between relations. Specifically, we present examples to show the the top three relevant relations of some relations in Table \ref{tab: top3-1}.
	
	\begin{table}[ht]
		\vspace{0mm}
		\caption{Some relations and their top 3 relevant relations. The relations are taken from WN18RR and NELL-995. We use CP to represent correlation pattern, and use CC to represent correlation coefficient.}
		\vspace{0mm}
		\resizebox{1.0\columnwidth}{!}{
			\begin{tabular}{c c c c}
				\toprule
				Target relation  &  Most relevant relations & CP & CC \\
				\midrule
				&\textit{\_hypernym}     & T-T       &0.71 \\
				\textit{\_member\_of\_domain\_usage} &\textit{\_similar\_to} & H-H  &0.50 \\
				&\textit{\_derivationally\_related\_form} & H-T     &0.42 \\
				\midrule  
				&\textit{city\_located\_in\_geopolitical\_location}   & H-H   &0.0843 \\
				\textit{person\_born\_in\_location} & \textit{organization\_terminated\_person} & T-T &0.0748\\
				&\textit{state\_located\_in\_country} & H-T           &0.0717 \\
				\midrule
				& \textit{hotel\_in\_city} & T-H & 0.1681 \\
				\textit{language\_of\_country} & \textit{museum\_in\_city} & H-H &0.0910 \\
				& \textit{league\_stadiums}       & H-T                 &0.0900 \\
				\bottomrule
			\end{tabular}
		}
		\label{tab: top3-1}
		\vspace{-3mm}
	\end{table}

	\begin{table*}[t]
		\caption{MRR and H@1 results on the inductive benchmark datasets extracted from WN18RR, FB15k-237, and NELL-995. We reimplement the three baselines Neural LP, DRUM, and GraIL under the inductive relation prediction protocol, with all the hyperparameters keeping the same with their original papers for a fair comparison. 
		}\label{Tab: MRR}
		\centering
		\resizebox{2.0\columnwidth}{!}{
			\begin{tabular}{l c c c c c  c c c c  c c c c } 
				\toprule
				&\multicolumn{4}{c}{\textbf{WN18RR}}&  \multicolumn{4}{c}{\textbf{FB15k-237}} & \multicolumn{4}{c}{\textbf{NELL-995}}\\
				\cmidrule(lr){2-5} \cmidrule(lr){6-9} \cmidrule(lr){10-13}
				&\multicolumn{2}{c}{v2}& \multicolumn{2}{c}{v3}& \multicolumn{2}{c}{v2}& \multicolumn{2}{c}{v3}& \multicolumn{2}{c}{v2}& \multicolumn{2}{c}{v3}\\
				\cmidrule(lr){2-3}\cmidrule(lr){4-5}
				\cmidrule(lr){6-7}\cmidrule(lr){8-9}
				\cmidrule(lr){10-11}\cmidrule(lr){12-13}
				& MRR   & H@1 & MRR   & H@1 & MRR   & H@1 & MRR   & H@1  & MRR   & H@1 & MRR   & H@1 \\
				\midrule
				{Neural LP} & .416 & .236 & .162 & .030 & .060 & .036 & .060 & .039 & .104 & .057 & .058 & .033 \\
				{DRUM}      & .216 & .034 & .247 & .141 & .052 & .034 & .047 & .027 & .101 & .055 & .090 & .038 \\
				{GraIL}     & .772 & .628 & .549 & .377 & .049 & .013 & .045 & .003 & .058 &	.010 & .042 &	.007 \\
				\midrule
				TACT-base     & .982 & .969 & \textbf{.913} & \textbf{.852} & .784 & .667 & \textbf{.827} & \textbf{.722} & .534 & .330 & .445 & .262 \\
				TACT    & \textbf{.984} & \textbf{.978} & .895 & \textbf{.852} & \textbf{.815} & \textbf{.717} & .752 & .609 & \textbf{.680} & \textbf{.533} & \textbf{.548} & \textbf{.354}  \\
				\bottomrule
			\end{tabular}
		}
	\end{table*}

	\begin{table*}[t]
		\caption{Further experiments for investigating the effect of each part of input embeddings in the scoring network of TACT. The symbols $\textbf{n}$, $\textbf{g}$, and $\textbf{r}$ represent the node embedding, the graph embedding, and the final relation embedding, respectively. We can set the selection of the embedding parts as a hyperparameter to get the best performance of TACT.}\label{Tab: ablation-2}
		\centering
		\resizebox{2.0\columnwidth}{!}{
			\begin{tabular}{c c c  c c c c c  c c c c  c c c c } 
				\toprule
				&&&\multicolumn{4}{c}{\textbf{WN18RR}}&  \multicolumn{4}{c}{\textbf{FB15k-237}} & \multicolumn{4}{c}{\textbf{NELL-995}}\\
				\cmidrule(lr){4-7} \cmidrule(lr){8-11} \cmidrule(lr){12-15}
				&&&\multicolumn{2}{c}{v1}& \multicolumn{2}{c}{v4}& \multicolumn{2}{c}{v1}& \multicolumn{2}{c}{v4}& \multicolumn{2}{c}{v1}& \multicolumn{2}{c}{v4}\\
				\cmidrule(lr){4-5}\cmidrule(lr){6-7}
				\cmidrule(lr){8-9}\cmidrule(lr){10-11}
				\cmidrule(lr){12-13}\cmidrule(lr){14-15}
				\textbf{n} & \textbf{g} & \textbf{r} & MRR   & H@1 & MRR   & H@1 & MRR   & H@1 & MRR   & H@1  & MRR   & H@1 & MRR   & H@1 \\
				\midrule
				&&\checkmark           & .990 & .983 & .981    & .966 & .804 & .700 & .593 & .410          & .877  & .756 & .304 & .171 \\
				&\checkmark&\checkmark & \textbf{.996} & \textbf{.996} & .987    & .978  & .824&.737 & \textbf{.652} & \textbf{.492} & .872  & .764 & .443 & .300 \\
				\checkmark&&\checkmark   & \textbf{.996} & .995 & \textbf{.992}    & \textbf{.989} & .806&	.702 & .570 & .380 & .819  & .696 & .399 & .254 \\ \midrule
				\checkmark&\checkmark&\checkmark& .995 & .995 & .988 & .982 & \textbf{.830} & \textbf{.741} & .575 & .378 & \textbf{.880} & \textbf{.776} & \textbf{.571} & \textbf{.444}  \\
				\bottomrule
			\end{tabular}
		}
	\end{table*}

\end{document}